\documentclass[letterpaper]{article} 
\usepackage{aaai25}  
\usepackage{times}  
\usepackage{helvet}  
\usepackage{courier}  
\usepackage[hyphens]{url}  
\usepackage{graphicx} 
\usepackage{natbib}  
\usepackage{caption} 
\usepackage{algorithm}
\usepackage{algorithmic}
\usepackage{booktabs}
\usepackage{amsmath} 
\usepackage[T1]{fontenc}
\usepackage[usenames,dvipsnames]{xcolor}
\usepackage{tikz}

\newcommand{\eg}{\textit{e}.\textit{g}., }

\usepackage{pgfplots}
\usetikzlibrary{patterns}
\pgfplotsset{compat=1.15}

\definecolor{babyblue}{rgb}{0.54, 0.81, 0.94}
\definecolor{babyred}{RGB}{252, 165, 165}

\usepackage{newfloat}
\usepackage{listings}
\DeclareCaptionStyle{ruled}{labelfont=normalfont,labelsep=colon,strut=off} 
\floatstyle{ruled}
\newfloat{listing}{tb}{lst}{}
\floatname{listing}{Listing}
\nocopyright 

\title{Irony in Emojis: A Comparative Study of Human and LLM Interpretation}
\author{
    Yawen Zheng\textsuperscript{\rm 1},
    Hanjia Lyu\textsuperscript{\rm 2},
    Jiebo Luo\textsuperscript{\rm 2}
}
\affiliations{
    \textsuperscript{\rm 1}The Chinese University of Hong Kong, Shenzhen\\
    \textsuperscript{\rm 2}University of Rochester\\
    121090840@link.cuhk.edu.cn,
    hlyu5@ur.rochester.edu,
    jluo@cs.rochester.edu%
}

\begin{document}

\maketitle

\begin{abstract}
Emojis have become a universal language in online communication, often carrying nuanced and context-dependent meanings. Among these, irony poses a significant challenge for Large Language Models (LLMs) due to its inherent incongruity between appearance and intent. This study examines the ability of {\tt GPT-4o} to interpret irony in emojis. By prompting {\tt GPT-4o} to evaluate the likelihood of specific emojis being used to express irony on social media and comparing its interpretations with human perceptions, we aim to bridge the gap between machine and human understanding. Our findings reveal nuanced insights into {\tt GPT-4o}’s interpretive capabilities, highlighting areas of alignment with and divergence from human behavior. Additionally, this research underscores the importance of demographic factors, such as age and gender, in shaping emoji interpretation and evaluates how these factors influence {\tt GPT-4o}’s performance. 
\end{abstract}

%

\section{Introduction}\label{sec:intro}

The emergence and advancement of Large Language Models (LLMs) have enabled more sophisticated simulations of human behavior, including the actions of social media users, through generative agents~\cite{park2023generative}. 
Among the diverse forms of online communication, emojis have evolved from simple pictorial representations to a universal language transcending cultural boundaries~\cite{2017nice, ai2017untangling}. Despite their widespread use, \citet{lyu2024human} highlighted notable discrepancies between the interpretation of emojis by {\tt GPT-4V} and human behavior, underscoring a critical gap in understanding.

Irony, a pervasive literary and communicative technique on social media, is defined by an inherent incongruity~\cite{zhang2019irony}. \citet{weissman2018strong} identified two primary carriers of irony on social media: verbal content and emojis. 
Emojis often embody irony through the contrast between their outward appearance and intended meaning, complicating their interpretation. Their euphemistic, humorous, and context-dependent uses further challenge the ability of LLMs to accurately discern sentiment~\cite{lyu2024human}. 
Addressing this challenge is essential, as accurate detection of irony in emojis could significantly enhance applications such as virtual assistants, chatbots, and sentiment analysis tools~\cite{10.1145/3709005}.

This study investigates the following research question:

\begin{itemize}
    \item How does {\tt GPT-4o}’s interpretation of ironic emojis compare to that of humans?
\end{itemize}
To answer this, we prompt {\tt GPT-4o} to assess how likely it is to choose a specific emoji to express irony on social media and compare its responses to human perceptions.

By focusing on irony in emojis, this research aims to evaluate LLMs' ability to understand subtle, sentiment-rich elements of this emerging universal language. Beyond technical insights, the study underscores the broader implications of AI in enhancing communication and human behavior simulation.

\section{Related Work}\label{sec:related_work}

The use of emojis on social media has attracted significant research attention, particularly in the context of sentiment analysis~\cite{hu2017spice}. The advent of large language models (LLMs) has introduced innovative approaches for analyzing sentiment involving emojis~\cite{wankhade2022survey, weissman2018strong}. However, a notable gap remains between LLMs' interpretation of emojis and human understanding. For instance, \citet{lyu2024human} identified significant discrepancies in behavior between humans and LLMs, which can be attributed to the subjective nature of emoji interpretation and the limitations imposed by cultural biases and insufficient representation of non-English cultures. Similarly, \citet{zhou2024emojis} demonstrated that while ChatGPT exhibits extensive knowledge of emojis, it also perpetuates stereotypes across communities. Furthermore, \citet{qiu2024semantics} highlighted that LLMs face challenges in suggesting emojis that align with the semantic meaning of social media posts. Building on this foundation, our study focuses on investigating how LLMs interpret emojis within a specific category: ironic emojis.

Several studies have explored factors influencing human interpretation of emojis, including personality traits~\cite{li2018mining} and demographics.
Gender and age are frequently cited as having a profound impact on emoji comprehension, with additional associations identified between cultural background, religious beliefs, and emoji interpretation~\cite{guntuku2019studying, wang2022sarcastic}. These demographic influences, particularly age and gender, are central to our investigation.

Gender-based differences in emoji comprehension have been well-documented. 
Differences are observed in accuracy in interpreting sentiments conveyed by emojis—such as joy, sorrow, fear, and anger—though no significant gender differences have been observed in the interpretation of surprise or disgust~\cite{chen2024individual}. 
Additionally, preferences for emojis that express positive sentiments differ across gender groups~\cite{chen2018through}. 
Age also plays a critical role in emoji usage and comprehension. Younger individuals often exhibit more advanced emoji usage skills, characterized by greater diversity, more nuanced applications, and higher accuracy in interpreting sarcasm and irony~\cite{chen2018through, garcia2022emoji, chen2024individual}.

Our study contributes to this body of work by conducting a fine-grained analysis of emoji interpretation through the lens of demographic factors. By incorporating age and gender information directly into prompts for LLMs, we aim to understand better how these models account for demographic variations in interpreting ironic emojis.

\section{Method}\label{sec:method}
This section describes the experimental procedure and its results. A quantitative analysis is conducted to evaluate the ability of the {\tt GPT-4o} variant to interpret irony.

\subsection{Human Perception of Emoji Irony}
We quantify how humans perceive the irony of emojis by analyzing their usage patterns. 
Specifically, we measure the frequency with which an emoji is used to convey irony in real-world social media posts and calculate its relative proportion of ironic usage as an \textbf{irony score}.
We follow \citet{xiang2020ciron} and define ``irony'' as instances where an emoji conveys a meaning opposite to its literal interpretation, resulting in a reversal of understanding.

We use the Ciron dataset compiled by \citet{xiang2020ciron}, which consists of over 8,700 posts (including around 3,000 posts containing emojis) from Weibo, a Chinese social media platform.
We select Ciron over the SemEval 2018 Irony Detection in English Tweets dataset~\cite{van-hee-etal-2018-semeval}, which, despite being an established English dataset, includes only 494 tweets containing emojis.

Each post is independently annotated by five postgraduate students, all native Chinese speakers, with an irony rating on a scale from 1 to 5, where 1 indicates ``not ironic," 2 ``unlikely ironic," 3 ``insufficient evidence of irony," 4 ``weakly ironic," and 5 ``strongly ironic."  
Table~\ref{tab:ciron_distribution} presents the distribution of irony scores across the collected posts.
See \citet{xiang2020ciron} for further annotation details. 

\begin{table}[h]
    \centering
    \begin{tabular}{lcc}
    \toprule[1.1pt]
    Category & Count & Percentage\\
    \midrule
     Not ironic    & 4,342 & 49.5\% \\
     Unlikely ironic    & 3,391 & 38.7\%\\
    Insufficient evidence of irony & 64 & 0.7\%\\
    Weakly ironic & 837 & 9.6\%\\
    Strongly ironic & 129 & 1.5\% \\
    \bottomrule[1.1pt]
    \end{tabular}
    \caption{Post irony distribution of Ciron~\cite{xiang2020ciron}.}
    \label{tab:ciron_distribution}
\end{table}

We use the irony rating of a post to represent the irony level of the emojis found within it. For instance, if a post is labeled as ``5 (strongly ironic),'' the emojis used in that post are considered strongly ironic. 
To compute the irony score $S(e)$ for a specific emoji $e$, we take the average of the irony scores of all posts $P_{e}$ that contain that emoji:
\begin{equation}
    S(e)=\frac{\sum_{p\in P_{e}}(R(p))}{|P_{e}|}
\end{equation}
where $R(p)$ represents the irony rating of post $p$, and $|P_{e}|$ denotes the total number of posts containing the emoji $e$. For instance, consider the ``kiss'' emoji \includegraphics[height=1em]{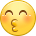}, which appears in 27 posts. Among these, 23 posts have an irony rating of 1, two posts have a rating of 2, and two posts have a rating of 3. Using the formula, the irony score of the ``kiss'' emoji is calculated as:
\begin{equation}
    S(\includegraphics[height=1em]{figures/kiss.png})=\frac{(23 \times 1) + (2 \times 2) + (2 \times 3)}{27}=1.22
\end{equation}

\subsection{GPT-4o's Classification of Emoji Irony}
To assess {\tt GPT-4o}'s classification of irony, we prompt it to evaluate how likely it would use a given emoji in a social media post to express irony. 

\paragraph{Prompt Design}

The exact prompt reads:
``{\it Imagine you are a social media user; rate your likelihood of using this emoji if your intention is to express irony on an 11-point scale (11=very likely, 1=very unlikely). The rating may depend on the context. You need to give the most likely rating and your explanation. Do not give multiple ratings in terms of scenarios. Only one rating is required.}''
Following the approach of \citet{contextfree} and \citet{lyu2024human}, we provide the model with emojis in image format.

We further investigate whether {\tt GPT-4o}'s classification changes when demographic information is included in the prompt by revising the first sentence to ``{\it Imagine you are a [gender] social media user aged [age] ...}''

The study considers five age groups: “under 20,” “20-34,” “35-49,” “50-64,” and “over 65.” 
Following previous studies~\cite{kotek2023gender,wan2023kelly}, gender groups include male and female, and experiments are conducted with all combinations of these demographic categories.
We recognize that gender identity is diverse and encompasses a wide spectrum, including non-binary and other gender non-conforming identities that are frequently underrepresented. The binary gender framework applied in this study does not capture the complexity of human gender expression.
Given these limitations, we approach the results with care, fully aware of the constraints inherent in the current dataset and methodology. We encourage future research to adopt more inclusive and representative frameworks.

\paragraph{Experiment Setting}

To ensure the robustness and consistency of the results, the model {\tt GPT-4o} is queried three times for each emoji used in the study to assess irony scores, with its temperature set to 0.5.

\section{Results}\label{sec:result}

The collected social media posts contain a total of 82 unique emojis. Table~\ref{tab:ciron_emoji_descriptive_stat} presents the descriptive statistics of human-perceived and model-classified irony scores for these emojis.
To enable comparison, the model's ratings are rescaled from a range of 1–11 to align with the 1–5 scale.
Using the Wilcoxon signed-rank test, we find that the median irony score assigned by {\tt GPT-4o} is significantly higher than the scores perceived by humans ($W=918.5$, $p<.001$). 
This indicates that, on average, {\tt GPT-4o} considers the same emoji more likely to be used for expressing irony compared to human perception.
This discrepancy may stem from {\tt GPT-4o} being training on data with a disproportionate representation of ironic emoji usage.
We explore this issue further in the following section.
The irony scores rated by {\tt GPT-4o} also exhibit greater variability.

\begin{table}[ht]
    \centering
    \setlength{\tabcolsep}{1.5mm}
    \begin{tabular}{cccccc}
    \toprule[1.1pt]
  &  Mean & Std & Min & Median & Max \\
    \midrule
Human-perceived    & 1.73 & 0.58 & 1.00 & 1.65 & 4.00 \\ 
Model-classified & 2.21 & 1.58 & 1.00 & 1.80 & 4.73 \\
    \bottomrule[1.1pt]
    \end{tabular}
    \caption{Descriptive statistics of human-perceived and normalized model-classified irony scores for emojis.}
    \label{tab:ciron_emoji_descriptive_stat}
\end{table}

\input{figures/fig_emoji_corr}

Moreover, as shown in Figure~\ref{fig:human_vs_model_corr}, by using the Spearman correlation test, we find that the irony scores assigned by {\tt GPT-4o} and those perceived by humans show a significant positive correlation ($\rho=0.28$, $p<.05$). 
Emojis located closer to the dashed line reflect a higher degree of alignment between {\tt GPT-4o}'s classification of their use for expressing irony and human perception.

To explore this alignment further, we prompt {\tt GPT-4o} to interpret emojis. For example, {\tt GPT-4o} explains that the emoji ``\includegraphics[height=1em]{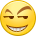}'' can convey irony due to its nuanced facial expression. The smirk's inherent ambiguity makes it well-suited for ironic statements, where the intended meaning diverges from the literal interpretation. This emoji can suggest sentiments like ``I know something you don't" or ``I'm not being entirely serious," which are consistent with the subtle and indirect nature of irony.

However, in the Ciron dataset, we observe that the usage of this emoji differs. For instance, one post using this emoji states, ``A Weibo post that challenges your psychological limits—dare to try?" This example highlights a distinct context where the emoji may not primarily signify irony.

Another emoji, ``\includegraphics[height=1em]{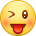},'' is rated highly by {\tt GPT-4o} for its association with irony but receives lower ratings from human perception. In Ciron, this emoji is frequently linked to positive and playful sentiment. When prompted, {\tt GPT-4o} explains that the emoji may also imply a teasing or mocking tone, suggesting an alternate interpretation that sometimes aligns with an ironic context.

\subsection{Prompts with Demographic Information}

\begin{figure}[t]
    \centering
    \begin{tikzpicture}
    \begin{axis}[
        width=\linewidth,
        height=6cm,
        ytick style={draw=none},
        ybar,
        xtick=data,
        symbolic x coords={\footnotesize{Under 20}, \footnotesize{20-34}, \footnotesize{35-49}, \footnotesize{50-64},\footnotesize{Over 65}},
        ylabel={Irony Score},
        xticklabel style={align=center},
        bar width=8pt,
        grid=major,
        enlarge x limits=0.15, 
        legend style={at={(0.5,1.15)}, anchor=north, legend columns=-1, draw=none}
      ]
      \addplot+ [fill=babyred, draw=black,postaction={pattern=north west lines}] coordinates {(\footnotesize{Under 20}, 2.33) (\footnotesize{20-34}, 2.30) (\footnotesize{35-49}, 2.15) (\footnotesize{50-64}, 1.98) (\footnotesize{Over 65}, 1.90)};
      \addplot [fill=babyblue, draw=black] coordinates {(\footnotesize{Under 20}, 2.37) (\footnotesize{20-34}, 2.27) (\footnotesize{35-49}, 2.14) (\footnotesize{50-64}, 2.01) (\footnotesize{Over 65}, 1.87)};
      \legend{Female, Male}
    \end{axis}
    \end{tikzpicture}
    \caption{
    When the prompt includes demographic information, no significant differences in irony scores are observed between prompts specifying female or male gender. However, the irony scores tend to decrease on average as the specified age in the prompt increases.}
    \label{fig:irony_emoji_bar}
\end{figure}
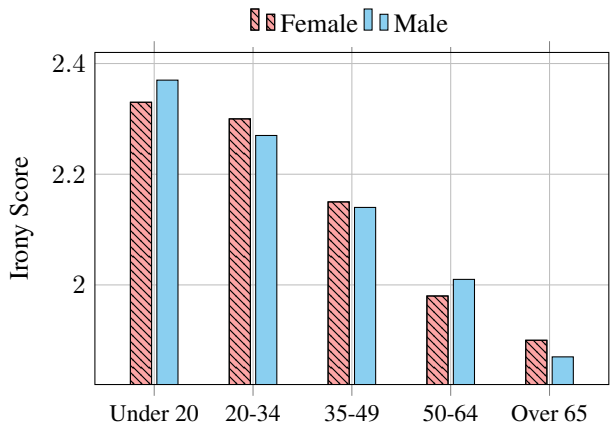

This section explores the classification of emoji irony by {\tt GPT-4o} when demographic information, including age and gender, is incorporated into the prompt.

Figure~\ref{fig:irony_emoji_bar} presents the average irony scores assigned by {\tt GPT-4o} to each emoji across prompts specifying different demographic attributes. The results suggest minimal differences in irony scores between prompts indicating female or male gender. However, a notable pattern emerges with age: irony scores tend to decrease as the specified age in the prompt increases.

The observed differences across age groups may be influenced by generational variations in communication habits, cultural norms, and digital literacy. Men and women of the same age group might exhibit similar interpretations of emoji irony due to shared digital environments, where gender differences in language use can be less pronounced. Shared cultural exposure to digital norms, memes, and emoji trends may contribute to this similarity.

In contrast, age appears to play a more substantial role in shaping perceptions of emoji irony. Younger individuals, having grown up with emojis as an integral part of their communication toolkit, may use and interpret emojis with greater fluidity and creativity. They are more likely to assign nuanced or ironic meanings to emojis compared to older individuals, who might favor more literal interpretations.

The meanings and contexts of emojis have evolved significantly over time, often driven by younger generations. For instance, emojis like “\includegraphics[height=1em]{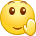}” or “\includegraphics[height=1em]{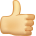}” might be perceived as straightforward by older users, while younger users may attribute layered, ironic, or sarcastic meanings to the same symbols.

Additionally, younger users may gravitate toward platforms like TikTok or Instagram, where ironic or sarcastic emoji use is more prevalent, whereas older users might primarily engage with platforms such as email or Facebook, where emoji use tends to be more literal. 

These insights align with findings from \citet{garcia2022emoji} and \citet{chen2024individual}, which indicate that younger individuals generally demonstrate a heightened ability to detect irony in emoji usage compared to older cohorts. While these results provide a compelling perspective, further research is necessary to better understand the interplay of demographic factors in shaping emoji interpretation.

\section{Discussions and Conclusions}\label{ssec:discussions_conclusions}

In this study, we investigate how {\tt GPT-4o} evaluates the level of irony in emojis compared to human perceptions. 

Our findings reveal that, on average, {\tt GPT-4o} assigns higher irony scores to emojis than humans do for the same emoji. This observation has several implications. First, {\tt GPT-4o} may have been trained on data with a disproportionate representation of ironic emoji usage, leading to an inflated assessment of their ironic potential. Second, the model might overgeneralize irony based on patterns in its training data, potentially lacking the contextual nuance humans rely on when interpreting emojis in specific scenarios. Lastly, this overestimation could have practical implications, such as misinterpretations in applications where accurate understanding of human communication patterns is critical—examples include sentiment analysis, chatbot interactions, and social media analysis.

We also observe minimal differences in irony scores between prompts indicating female or male genders. However, a notable trend emerged with age: irony scores tend to decrease as the specified age in the prompt increases. This age-related pattern aligns with findings from prior studies.

An additional factor contributing to the discrepancy between model-classified and human-perceived scores could be the dataset used for evaluation. We conduct the study using a Chinese social media post dataset, while {\tt GPT-4o} is predominantly trained on English-language data. This language orientation, combined with cultural differences in the use and interpretation of emojis, may partially explain the gap. However, due to the unavailability of an English-language irony dataset, we are unable to evaluate {\tt GPT-4o}’s performance on English-language emojis. Addressing this limitation is a key focus of our future work.

Another limitation of our study is that it focuses exclusively on one model, {\tt GPT-4o}. To gain a more comprehensive understanding of irony evaluation by large language models, future work will include a broader range of LLMs, especially those not primarily oriented toward English. This will allow for a more robust analysis of how linguistic and cultural differences impact irony perception.

\bibliography{aaai25}

\appendix
\section{Appendix}
\subsection{Further Discussion on Potential Broader Impact and Ethical Considerations}
 By improving LLMs’ understanding of ironic emojis, our work can enhance human-machine interactions in applications such as virtual assistants, chatbots, and sentiment analysis systems. These improvements can lead to more effective and context-aware communication tools.
 Incorporating demographic information (\eg age, gender) into AI models raises concerns about perpetuating or amplifying stereotypes. For example, demographic-based prompts might unintentionally reinforce assumptions about how certain groups use emojis. Care must be taken to ensure that the models remain equitable and do not propagate biases.

\end{document}